%% file: main.tex
\documentclass{article} 
\usepackage{iclr2027_conference,times}

\input{math_commands.tex}

\usepackage{hyperref}
\usepackage{url}

\usepackage{amssymb}
\usepackage{graphicx}
\usepackage{booktabs}     
\usepackage{tabularx}     
\usepackage{wrapfig}      
\usepackage{multirow}
\usepackage{fontawesome}
\usepackage{xcolor}
\usepackage{colortbl}     
\usepackage{xspace}       
\usepackage{float}        
\usepackage{titletoc}     
\usepackage[many]{tcolorbox}  

\input{macros.tex}

\title{RouteWeaver: Weaving Mode Selection and Execution into Unified LLM Routing}

\author{
Xiaohan Wang$^{1,2}$\thanks{Work done during an internship at Nanyang Technological University.} \quad
Haozhen Zhang$^{1}$ \quad
Qingyuan Liu$^{1}$ \quad
Tao Feng$^{3}$ \quad
Wenya Wang$^{1}$ \\
$^{1}$Nanyang Technological University \\
$^{2}$Huazhong University of Science and Technology \\
$^{3}$University of Illinois Urbana-Champaign \\
\texttt{shawn\_wang@hust.edu.cn, wangwy@ntu.edu.sg}
}

\iclrfinalcopy
\begin{document}

\maketitle

\begin{abstract}
\input{sections/abstract}

\begin{center}
\faGithub\ \href{https://github.com/LaughKing/RouteWeaver}{\texttt{RouteWeaver}} \quad
\faGlobe\ \href{https://laughking.github.io/RouteWeaver/}{\texttt{Project Website}} \quad
\raisebox{-0.2ex}{\includegraphics[height=1em]{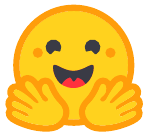}}\ \href{https://huggingface.co/collections/e2rea1/routeweaver}{\texttt{Hugging Face}}
\end{center}

\end{abstract}

\input{sections/intro}

\input{sections/related}
\input{sections/method}
\input{sections/experiments}

\input{sections/conclusion}


\input{sections/statements}

\bibliography{main}
\bibliographystyle{iclr2027_conference}

\appendix
\input{sections/appendix}

\end{document}

%% file: math_commands.tex
\usepackage{amsmath,amsfonts,bm}

\def\eqref#1{equation~\ref{#1}}

\def\1{\bm{1}}

\DeclareMathAlphabet{\mathsfit}{\encodingdefault}{\sfdefault}{m}{sl}
\SetMathAlphabet{\mathsfit}{bold}{\encodingdefault}{\sfdefault}{bx}{n}



%% file: macros.tex
\definecolor{oursrow}{RGB}{226,237,252}
\definecolor{impblue}{RGB}{47,79,168}
\newcommand{\imp}[1]{\,{\scriptsize\textcolor{impblue}{(+#1\%)}}}
\definecolor{dropgray}{RGB}{110,110,110}

\newcommand{\ours}{\texttt{RouteWeaver}\xspace}



%% file: sections/abstract.tex

LLM routing dynamically selects and coordinates heterogeneous language models across diverse tasks. However, existing approaches are largely built around a predefined execution paradigm, such as single-round, multi-round, or agentic routing, which fixes how model calls are organized and limits adaptation to different query needs. We propose \ours, a unified routing framework that integrates these routing modes within a single LLM policy, jointly learning which mode to use and how to select models within that mode. To jointly optimize mode selection and within-mode execution, we develop COMET-GRPO (\textbf{CO}ordinated \textbf{M}ode--\textbf{E}xecution \textbf{T}raining), which decouples the trajectory-level advantage into separate signals for mode selection and within-mode execution, enabling more targeted credit assignment to each decision level. We further introduce progressive exploration, which gradually shifts training from externally assigned modes to autonomous selection, so that the router first learns how to execute each mode before deciding when to use it. We evaluate \ours across mathematics, coding, knowledge, logical reasoning, and factual recall. \ours achieves an average accuracy of 61.4\%, an 11.03\% relative improvement over the strongest baseline, and outperforms all compared baselines in each of the five domains. Ablations show that both COMET-GRPO and progressive exploration are important for adaptive routing; without progressive exploration, the routing policy collapses entirely to single-round execution.

%% file: sections/intro.tex

\section{Introduction}
\label{sec:intro}

The growing diversity of large language models (LLMs) has motivated LLM routing as a paradigm for leveraging complementary capabilities, where different models are selected and coordinated according to their strengths, costs, and suitability for each query~\citep{DBLP:journals/tmlr/ChenZ024, DBLP:conf/iclr/DingM0SMRLA24, hu2024routerbench, DBLP:conf/iclr/OngAWC0GKS25}. Depending on how model calls are organized, current routing approaches generally follow three paradigms: single-round routing selects a single model for one-shot execution~\citep{chen2024routerdc, feng2025graphrouter, wang2026iclrouter}, multi-round routing makes successive model selections conditioned on intermediate outputs~\citep{zhang2025routerr1, shao2026r2reasoner}, and agentic routing organizes multiple models into structured collaborative workflows~\citep{yue2025masrouter, feng2026graphplanner}. Although routing decisions within each paradigm can adapt to the query, the execution paradigm itself is typically fixed in advance. As a result, the router can adapt \emph{which models} to use, but not \emph{how model calls should be organized} for different queries. This raises a natural question: \emph{how can we unify these routing paradigms within a single policy that jointly adapts execution mode and model selection to each query?}

Unifying these routing paradigms within a single policy is non-trivial. Single-round, multi-round, and agentic routing follow different execution structures and decision processes, requiring a common formulation that accommodates all three while preserving their distinct behaviors. Beyond this structural challenge, two learning problems are particularly important. \emph{First, different queries may favor different execution modes, yet determining which mode is appropriate requires reasoning about the task before its execution unfolds.} A unified router must therefore learn when single-round, multi-round, or agentic execution is most suitable for each query. \emph{Second, selecting an appropriate mode alone is insufficient, since each mode demands a different form of model selection and coordination.} Single-round routing requires choosing an effective model directly, multi-round routing must adapt subsequent selections to intermediate outputs, and agentic routing additionally organizes models through roles and dependencies. The router must therefore learn not only \emph{which mode to use}, but also \emph{how to execute effectively within the selected mode}.

To address these challenges, we propose \ours, a unified routing policy that learns to select among single-round, multi-round, and agentic execution paradigms and perform model selection within the chosen paradigm. We first introduce a unified routing format that expresses all three paradigms through a common action grammar for mode selection and subsequent routing decisions. To jointly optimize these two levels of decisions, we develop COMET-GRPO (\textbf{CO}ordinated \textbf{M}ode--\textbf{E}xecution \textbf{T}raining), which decouples the trajectory-level advantage into separate signals for mode selection and within-mode execution, enabling targeted credit assignment to each decision level. We further introduce progressive exploration, which gradually shifts training from externally assigned modes to autonomous selection, allowing the router to first learn how to execute within each mode before deciding when to use it.

We evaluate \ours on benchmarks spanning mathematics, coding, knowledge, logical reasoning, and factual recall. \ours achieves 61.4\% average accuracy, an 11.03\% relative improvement over the strongest baseline, and leads all baselines across the five domains. We further study performance--cost trade-offs and generalization on held-out benchmarks (AIME, GPQA-D, and BBH). With $\alpha=0.1$, Worker inference cost decreases by 23\% with only a 1.0-point accuracy drop, while \ours achieves 53.7\% average accuracy on held-out benchmarks, an 18.28\% relative improvement over the strongest baseline. Ablations and routing analyses further validate COMET-GRPO and progressive exploration and reveal adaptive execution choices across tasks.
Our main contributions are summarized as follows:
\begin{itemize}
    \item We formulate unified LLM routing across single-round, multi-round, and agentic execution paradigms, enabling a single policy to adapt both the execution mode and model selection to each query rather than committing to a predefined routing schema.

    \item We propose \ours, featuring COMET-GRPO, which decouples credit assignment for mode selection and within-mode execution, together with progressive exploration that gradually transitions the router from learning how to execute each mode to autonomously deciding when to use it for different queries.

    \item Extensive experiments across mathematics, coding, knowledge, logical reasoning, and factual recall demonstrate the effectiveness of unified routing over existing routers and fixed-mode strategies. Further analyses examine performance--cost trade-offs, routing behavior, and the contributions of the proposed training components.
\end{itemize}

%% file: sections/related.tex

\section{Related Work}
\label{sec:related}

\subsection{Execution Paradigms for LLM Routing}

Existing LLM routing methods can be categorized by how model calls are organized.
(1) \textbf{Single-round routing} selects one model for each query, typically based on model capability, query difficulty, or performance--cost trade-offs. RouteLLM~\citep{DBLP:conf/iclr/OngAWC0GKS25} learns routing from preference data, while RouterDC~\citep{chen2024routerdc} and GraphRouter~\citep{feng2025graphrouter} model query--model relations. ICL-Router~\citep{wang2026iclrouter} further improves routing through learned model representations.
(2) \textbf{Multi-round routing} allows routing decisions to evolve across calls using intermediate outputs. Router-R1~\citep{zhang2025routerr1} formulates routing as a sequential decision process with repeated model selection and response aggregation, while R2-Reasoner~\citep{shao2026r2reasoner} combines task decomposition with model allocation for intermediate reasoning.
(3) \textbf{Agentic routing} organizes multiple models into collaborative workflows. MasRouter~\citep{yue2025masrouter} jointly considers collaboration structures, agent roles, and model choices, while GraphPlanner~\citep{feng2026graphplanner} generates query-dependent workflows with model--role assignments and explicit interactions.
Despite adaptive routing within each paradigm, prior approaches fix the execution paradigm in advance. In contrast, \ours unifies single-round, multi-round, and agentic routing as selectable modes of one policy, jointly learning mode selection and within-mode execution.

\subsection{Reinforcement Learning for Agentic LLMs}

Reinforcement learning (RL) has become a central approach for improving the reasoning and interactive capabilities of LLMs.
GRPO~\citep{shao2024deepseekmath} replaces critic-based advantage estimation with group-relative rewards and has inspired several variants.
Dr.GRPO~\citep{DBLP:journals/corr/abs-2503-20783} addresses optimization biases in GRPO, while DAPO~\citep{DBLP:conf/nips/YuZZYZYDFLLLLLL25} improves training stability through asymmetric clipping and dynamic sampling.
GDPO~\citep{liu2026gdpo} further decouples reward normalization for multi-reward optimization.
RL has also been extended to interactive agents. Search-R1~\citep{DBLP:journals/corr/abs-2503-09516} and ReTool~\citep{DBLP:journals/corr/abs-2504-11536} train LLMs to interleave reasoning with search or tool interactions, while GiGPO~\citep{DBLP:conf/nips/FengXLA25} introduces episode- and step-level relative advantages for finer credit assignment in multi-turn agents.
Building on these advances, COMET-GRPO adapts group-relative optimization to unified LLM routing, where mode selection and within-mode execution form two coupled decision levels and receive separate credit signals.

%% file: sections/method.tex

\section{Preliminaries}
\label{sec:formulation}

\begin{figure}[t]
\centering
\includegraphics[width=\linewidth,trim=3.5 4 4 4,clip]{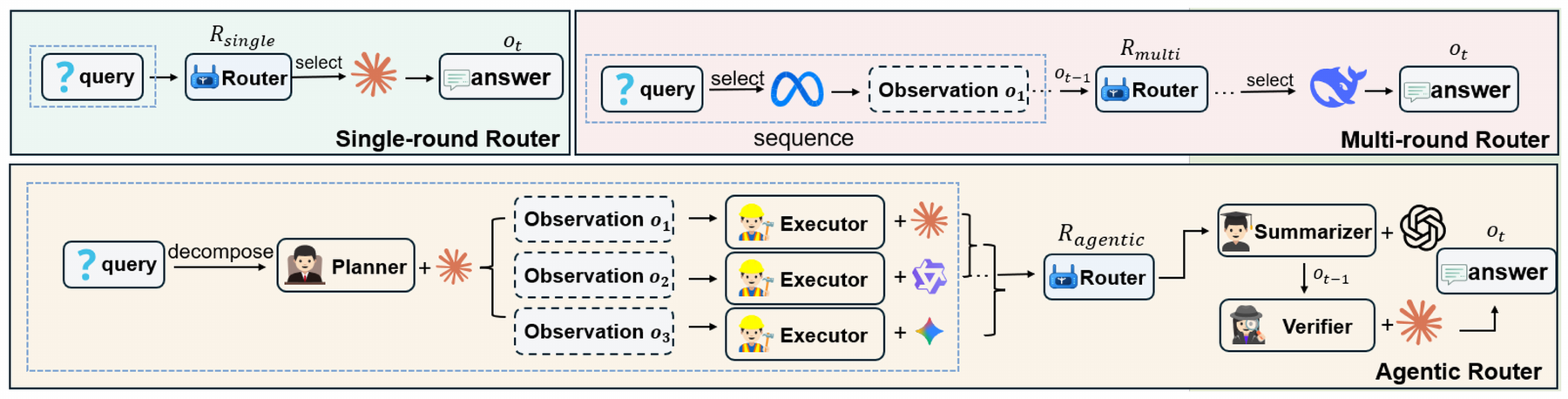}
\caption{
Three execution paradigms.
\textbf{Top left, single-round:} the router selects one Worker for a single call. 
\textbf{Top right, multi-round:} the router selects Workers sequentially using accumulated observations. 
\textbf{Bottom, agentic:} the router selects Workers, assigns roles, and specifies dependencies between calls; the depicted workflow is illustrative, and its roles and dependencies vary across queries.
}
\label{fig:paradigms}
\end{figure}

Given a query $x$ and a pool of Worker models
$\mathcal{W} = \{w_1,\ldots,w_N\}$, a router $\mathcal{R}$ determines which Workers to invoke and
how their calls are organized. We refer to the resulting pattern of Worker
invocation and information flow as the \emph{execution structure}. We distinguish
three execution paradigms according to how routing decisions are structured
(Figure~\ref{fig:paradigms}).

\textbf{Single-round routing.}
The router selects one Worker for the query and calls it once:
\begin{equation}
    w = \mathcal{R}_{\mathrm{single}}(x),
    \qquad
    o = w(x).
\end{equation}
Here, $w \in \mathcal{W}$ is the selected Worker and $o$ is its output. The resulting
execution structure consists of a single Worker call: the routing decision precedes
execution, with no subsequent Worker selection based on the returned response.

\textbf{Multi-round routing.}
The router makes successive calls using earlier feedback. Let
$s_t$ denote the sub-query generated for the $t$-th Worker call and
$h_{t-1} = \big((w_1,s_1,o_1),\ldots,(w_{t-1},s_{t-1},o_{t-1})\big)$ the ordered
history of preceding interactions. Each routing decision specifies a Worker and its
sub-query:
\begin{equation}
    (w_t,s_t) = \mathcal{R}_{\mathrm{multi}}(x,h_{t-1}),
    \qquad
    o_t = w_t(s_t).
\end{equation}
The response is appended to the history before the next decision. This sequential
execution structure allows intermediate outputs to guide model selection, task
refinement, and further reasoning.

\textbf{Agentic routing.}
Calls are organized into a workflow with explicit roles and dependencies. We
describe the completed workflow after $t$ calls by $G_t = (V_t,E_t)$, where $V_t$
indexes completed Worker calls and an edge $(j,k) \in E_t$ indicates that call $k$
uses the output of call $j$. A routing decision is
\begin{equation}
    (w_t,s_t,r_t,P_t) = \mathcal{R}_{\mathrm{agentic}}(x,G_{t-1}),
\end{equation}
where $r_t$ specifies the role, such as planning, execution, verification, or
summarization, and $P_t \subseteq V_{t-1}$ is the set of completed predecessor calls
selected as dependencies. The outputs $\{o_j : j \in P_t\}$ are supplied as context to
Worker $w_t$, which executes sub-query $s_t$ under role $r_t$, and its output is
recorded in the workflow. By selecting $P_t$, the router determines the direct
dependencies of each call. The resulting dependency structure supports branching and
aggregation, while independent calls may execute in parallel. Agentic workflows can
also be sequential; their defining distinction here is the explicit organization of
roles and dependencies.

\section{\texorpdfstring{\ours}{RouteFM}}
\label{sec:method}

Building on the above formulation, \ours unifies single-round, multi-round, and agentic routing in one policy. This raises three challenges. First, their heterogeneous execution structures require a common representation. Second, performance depends jointly on mode selection and within-mode execution, calling for differentiated learning signals. Third, mode selection requires sufficient experience across modes to avoid early bias. \ours addresses these challenges through a unified routing formulation (Section~\ref{sec:unified-router}), COMET-GRPO for coordinated mode--execution learning (Section~\ref{sec:mode-grpo}), and progressive exploration toward autonomous routing (Section~\ref{sec:progressive}).

\begin{figure}[!ht]
\centering
\includegraphics[width=\linewidth,trim=1.9 26.4 56.9 0,clip]{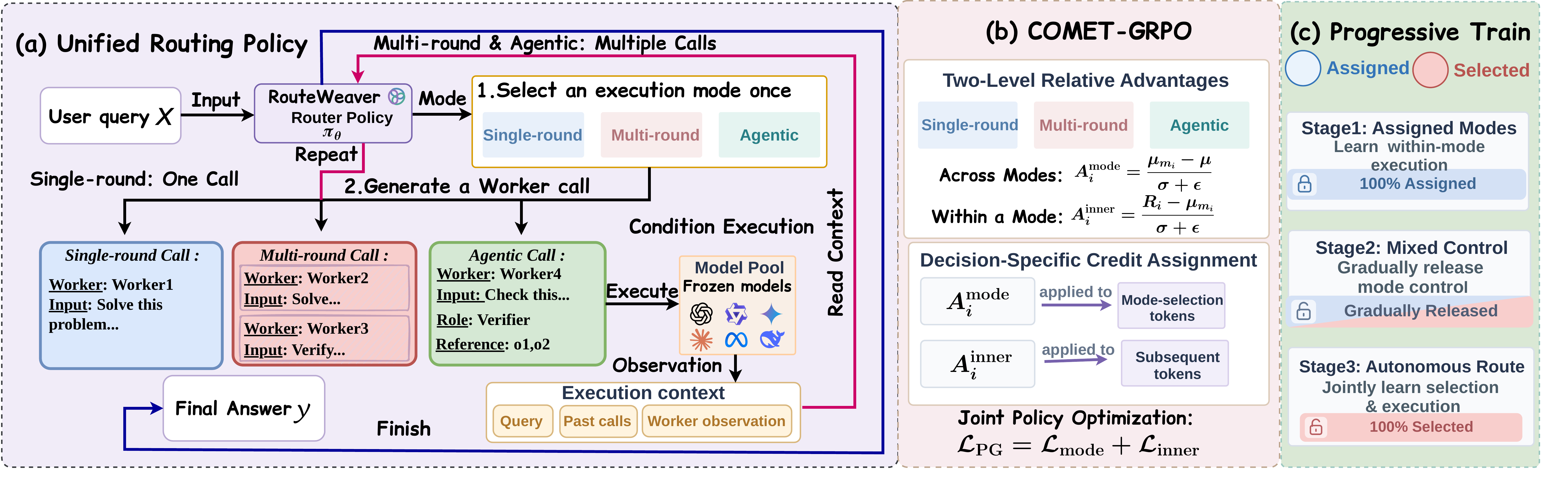}
\caption{Overview of \ours. \textbf{(a)} One routing policy $\pi_\theta$ selects an execution mode, then calls frozen Workers;Multi-round calls automatically use the latest observation, while agentic calls include role assignments and router-selected references to earlier outputs.Returned observations are appended to its context until it emits the final answer. 
\textbf{(b)}  COMET-GRPO computes mode-level and within-mode advantages per reward component, mixes them (Eq. (7)), and assigns $A_i^{\mathrm{mode}}$ to mode-selection tokens and $A_i^{\mathrm{inner}}$ to subsequent router tokens.
\textbf{(c)} Progressive training moves mode selection from assigned modes to the router in three stages.}
\label{fig:framework}
\end{figure}

\subsection{Unified Trajectory Representation and Execution}
\label{sec:unified-router}

To this end, we represent each routing process as a mode-conditioned trajectory.
Given a query $x$, the router first selects an execution mode
\[
    m \sim \pi_\theta(\cdot \mid x), \qquad
    m \in \mathcal{M} = \{\mathrm{single},\mathrm{multi},\mathrm{agentic}\},
\]
and subsequently generates routing actions conditioned on the query and the
accumulated interaction context. Each Worker call specifies a Worker $w$ and a
sub-query $s$; agentic calls additionally specify a role $r$ and references $P$ to
completed calls. Multi-round execution passes the previous observation to each
subsequent call, while agentic execution passes the referenced outputs to
role-structured and potentially parallel calls. Single-round execution contains
exactly one Worker call. The serialization format and mode-specific constraints are
detailed in Appendix~\ref{app:action-format}.

Executing an action returns the Worker outputs as an observation $o_t$, which is
appended to the router context before the next decision. The router continues issuing
mode-compatible actions until it produces the final answer $y$, yielding
\begin{equation}
    \tau = (m,a_1,o_1,\ldots,a_T,o_T,y).
    \label{eq:trajectory}
\end{equation}
The mode, routing actions, and final answer are generated by the router and optimized
as policy tokens, whereas Worker observations and other environment-inserted content
are used only as context.

Each trajectory receives a task reward and an efficiency reward,
\begin{equation}
    (r_{\mathrm{task}},r_{\mathrm{eff}}) =
    \begin{cases}
        (1,\,S_{\mathrm{cost}}(\tau)), & \tau \text{ is valid and correct},\\[2pt]
        (0,\,0), & \tau \text{ is valid but incorrect},\\[2pt]
        (-1,\,-1), & \tau \text{ is terminal-invalid},
    \end{cases}
    \qquad
    S_{\mathrm{cost}}(\tau) = \frac{1}{1+C(\tau)/C_{\mathrm{ref}}}.
    \label{eq:reward}
\end{equation}
Correctness is determined by the task-specific scorer (Appendix~\ref{app:eval}),
while $C(\tau)$ measures the fixed-reference inference cost of Worker calls, excluding
router computation, and $C_{\mathrm{ref}} > 0$ is a frozen reference cost
(Appendix~\ref{app:training}). Terminal-invalid trajectories arise from grammar or
execution-constraint violations. These trajectory-level signals jointly reflect mode
selection and subsequent execution, motivating the decision-specific credit
assignment introduced next.

\subsection{COMET-GRPO}
\label{sec:mode-grpo}

Joint mode selection and execution create an ambiguous credit-assignment problem. A
low-reward trajectory may result from selecting an unsuitable execution mode, from
executing an otherwise suitable mode poorly, or from both. Standard
GRPO~\citep{shao2024deepseekmath} does not
distinguish these cases: it assigns the same trajectory-level relative advantage to the
mode-selection tokens and all subsequent router-generated tokens. We therefore
decompose relative feedback into two complementary comparisons: an across-mode
comparison for mode selection and a within-mode comparison for execution conditioned
on the selected mode. COMET-GRPO (\emph{COordinated Mode--Execution Training}) uses
these two signals to optimize the two decision levels separately within a shared
policy.

\textbf{Two levels of relative advantage.}
For each query, we sample a group of $K$ rollouts; all statistics below are computed
within this group and separately for each reward component. Let $\mathcal{V}$ be the
rollouts retained for advantage estimation (Appendix~\ref{app:training}). For one
component with values $r_i$, let $\mu$ and $\sigma$ be the mean and standard deviation
of $\{r_i\}_{i \in \mathcal{V}}$, so that GRPO assigns
$A_i^{\mathrm{GRPO}} = (r_i-\mu)/(\sigma+\epsilon)$ to every router token of $\tau_i$.
In a \emph{free} group, each rollout selects its own mode, so the group need not
cover all modes. Let $\mathcal{V}_m \subseteq \mathcal{V}$ contain the retained
rollouts that selected $m$, with mean $\mu_m$ of the same component. For rollout $i$
with mode $m_i$,
\begin{equation}
    A_i^{\mathrm{mode}} = \frac{\mu_{m_i}-\mu}{\sigma+\epsilon},
    \qquad
    A_i^{\mathrm{inner}} = \frac{r_i-\mu_{m_i}}{\sigma+\epsilon}.
    \label{eq:advantages}
\end{equation}
Intuitively, $A_i^{\mathrm{mode}}$ measures how the average reward of the selected mode
compares with the overall rollout-group average for the same query, whereas
$A_i^{\mathrm{inner}}$ measures how this particular trajectory compares with the
average trajectory under the same mode. Both use the group standard deviation $\sigma$ rather than a within-mode
one, keeping the two signals on a common scale.

Following the reward-wise normalization of GDPO~\citep{liu2026gdpo}, we compute these
advantages separately for the task and efficiency rewards and then mix them:
\begin{equation}
    A_i^{l} = (1-\alpha)\,A_{i,\mathrm{task}}^{l} + \alpha\,A_{i,\mathrm{eff}}^{l},
    \qquad l \in \{\mathrm{mode},\mathrm{inner}\}.
    \label{eq:adv-mix}
\end{equation}
With $\alpha = 0$ only the task signal is used; $\alpha > 0$ adds the efficiency
signal. The weight $\alpha$ is fixed within a run and is independent of the
exploration schedule of Section~\ref{sec:progressive}, and no further batch-level
normalization is applied. Component-wise definitions are given in
Appendix~\ref{app:optim}.

\textbf{Decision-specific credit assignment.}
Having separated the two sources of feedback, we apply each signal only to the
decisions it is intended to supervise. The mixed $A_i^{\mathrm{mode}}$ is assigned to the tokens of the mode decision,
$S_i^{\mathrm{mode}}$, and $A_i^{\mathrm{inner}}$ to the subsequent router tokens,
$S_i^{\mathrm{inner}}$: Worker choices, roles, references, sub-queries, and the final
answer. Observations, injected content, and padding are masked. Following the PPO
clipped surrogate objective~\citep{schulman2017ppo}, for a router token
$z_{ik}$ with context $H_{ik}$ (indexed by $k$ to distinguish tokens from interaction
steps), the clipped token loss is
\begin{equation}
    \ell_{ik}(A) = -\min\!\big[r_{ik}A,\;
    \operatorname{clip}(r_{ik},1-\eta_-,1+\eta_+)A\big],
    \qquad
    r_{ik} = \frac{\pi_\theta(z_{ik} \mid H_{ik})}{\pi_{\mathrm{old}}(z_{ik} \mid H_{ik})},
    \label{eq:token-loss}
\end{equation}
with dual clipping for negative advantages as specified in
Appendix~\ref{app:optim}.

\textbf{Joint policy optimization.}
Each of the two losses is normalized by its own global count of effective tokens in the
mini-batch, and the results are added:
\begin{equation}
    \mathcal{L}_{\mathrm{PG}} =
    \frac{\sum_i \sum_{k \in S_i^{\mathrm{mode}}} \ell_{ik}(A_i^{\mathrm{mode}})}
         {N_{\mathrm{mode}}}
    +
    \frac{\sum_i \sum_{k \in S_i^{\mathrm{inner}}} \ell_{ik}(A_i^{\mathrm{inner}})}
         {N_{\mathrm{inner}}},
    \label{eq:pg-loss}
\end{equation}
where $N_{\mathrm{mode}}$ and $N_{\mathrm{inner}}$ count the effective mode and inner
tokens in the mini-batch, and an empty term is skipped. Mode tokens are typically far
fewer than execution tokens, so averaging over all tokens together would give the
mode term a small aggregate weight merely because it is short; separate normalization
removes this dependence on length. The policy is regularized toward the frozen initial router
$\pi_{\mathrm{ref}}$, distinct from $\pi_{\mathrm{old}}$, using a KL penalty
with coefficient $\beta$:
\begin{equation}
    \mathcal{L}_{\mathrm{KL}} =
    \frac{1}{N_{\mathrm{KL}}} \sum_{(i,k) \in \Omega_{\mathrm{KL}}} \kappa_{ik},
    \label{eq:kl}
\end{equation}
where $\Omega_{\mathrm{KL}}$ is the full effective policy mask, $N_{\mathrm{KL}}$ its
token count, and $\kappa_{ik}$ a per-token estimate of the KL divergence to
$\pi_{\mathrm{ref}}$. The estimator,
masks, and degenerate groups are described in Appendix~\ref{app:optim}.

\subsection{Progressive Exploration}
\label{sec:progressive}

If the router chose modes autonomously from the start, early preferences could reduce
the training opportunities of the other modes. Progressive exploration therefore
first provides execution experience within each mode and then gradually hands mode
selection to the router, in three phases indexed by the data-batch index $b$. Each query is \emph{forced} with
probability
\begin{equation}
    \rho_b =
    \begin{cases}
        1, & b < b_0,\\[2pt]
        1-\dfrac{b-b_0}{b_1-b_0}, & b_0 \le b < b_1,\\[6pt]
        0, & b \ge b_1,
    \end{cases}
    \label{eq:anneal}
\end{equation}
and otherwise \emph{free}; all $K$ rollouts of a query share this condition. A forced
query receives one externally assigned mode, with modes distributed across the batch,
and its injected mode prefix is masked.

\textbf{Cold start ($b < b_0$).}
This phase teaches execution within each mode before the router selects modes. All
queries are forced. We denote by $\mathcal{L}_{\mathrm{GRPO}}^{\mathrm{global}}$ the
ordinary GRPO loss, obtained by averaging $\ell_{ik}(A_i^{\mathrm{GRPO}})$ over all
effective router tokens (Eq.~\ref{eq:grpo-global}). The router is updated with
$\mathcal{L}_{\mathrm{GRPO}}^{\mathrm{global}}$ on the task reward alone; advantage
decomposition and the support term are inactive. All runs start independently from
the same initial router at step 0; after this phase each run continues its own
training trajectory with its fixed $\alpha$ (Appendix~\ref{app:impl}).

\textbf{Forced/free transition ($b_0 \le b < b_1$).}
Forced groups decrease and free groups increase. Free groups use the advantages of
Eq.~\ref{eq:adv-mix}. Forced groups compute $A_i^{\mathrm{GRPO}}$ for each enabled
reward component, mix them with the same weights, and apply the result to their
execution tokens, which enter the inner term of Eq.~\ref{eq:pg-loss}; their injected
mode tokens remain masked (Eq.~\ref{eq:mixed-adv}). To keep probability support on
every mode and prevent it from concentrating prematurely, we add a one-sided penalty
on $\bar{p}_m$, the normalized score of candidate mode $m$ averaged over the unique
queries in the batch (Appendix~\ref{app:optim}):
\begin{equation}
    \mathcal{L}_{\mathrm{support}} =
    \frac{1}{|\mathcal{M}|} \sum_{m \in \mathcal{M}}
    \max\!\left(0,\log \delta - \log(\bar{p}_m+\epsilon)\right).
    \label{eq:support}
\end{equation}
Because it acts on batch-level averages, individual queries can still favor different
modes. Its coefficient $\lambda_b = \lambda_0 \rho_b$ applies only during the
transition; the term is inactive in the other phases.

\textbf{Fully autonomous routing ($b \ge b_1$).}
All queries are free, and the router jointly learns mode selection and execution with
COMET-GRPO.

\textbf{Phase objectives.}
Across the three phases, the router minimizes
\begin{equation}
    \mathcal{L}_b =
    \begin{cases}
        \mathcal{L}_{\mathrm{GRPO}}^{\mathrm{global}} + \beta\,\mathcal{L}_{\mathrm{KL}},
        & b < b_0,\\[2pt]
        \mathcal{L}_{\mathrm{PG}} + \beta\,\mathcal{L}_{\mathrm{KL}}
        + \lambda_b \mathcal{L}_{\mathrm{support}},
        & b_0 \le b < b_1,\\[2pt]
        \mathcal{L}_{\mathrm{PG}} + \beta\,\mathcal{L}_{\mathrm{KL}},
        & b \ge b_1,
    \end{cases}
    \label{eq:phase-objective}
\end{equation}
where $\mathcal{L}_{\mathrm{GRPO}}^{\mathrm{global}}$ uses the task-only GRPO advantage
and $\mathcal{L}_{\mathrm{PG}}$ uses the $\alpha$-mixed advantages, including the mixed
GRPO advantages of forced groups in its inner term. Phase boundaries and coefficients
are reported in Appendix~\ref{app:impl}.

%% file: sections/experiments.tex

\section{Experiments}
\label{sec:experiments}

We evaluate \ours across five task domains to examine whether jointly selecting and
executing different routing modes improves over methods restricted to a fixed
execution paradigm. We further study its performance--cost trade-off, isolate the
contributions of COMET-GRPO and progressive exploration, and analyze how
execution-mode choices vary across tasks. Additional dataset, evaluation, and
implementation details are provided in the appendices.

\subsection{Experimental Setup}
\label{sec:exp-setup}

\textbf{Datasets.}
Our evaluation comprises nine benchmarks spanning five domains: \textbf{Math},
MATH~\citep{hendrycks2021math} and Omni-MATH~\citep{gao2025omnimath}; \textbf{Code},
APPS~\citep{hendrycks2021apps}, LiveCodeBench~\citep{jain2025livecodebench}, and
TACO~\citep{li2023taco}; \textbf{Knowledge}, \mbox{SuperGPQA}~\citep{du2025supergpqa};
\textbf{Reason}, SATBench~\citep{wei2025satbench}; and \textbf{Recall},
TriviaQA~\citep{joshi2017triviaqa} and
SimpleQA-Verified~\citep{haas2025simpleqaverified}. Test questions do not overlap with
the training questions used for router optimization. Dataset composition, training
pools, and evaluation sets are provided in Appendix~\ref{app:data}.

\textbf{Router and Worker models.}
The router is Qwen3-4B-Instruct-2507~\citep{yang2025qwen3} with all parameters
fine-tuned. The Worker pool
contains six frozen models: Qwen3-8B and
Qwen3-30B-A3B-Instruct-2507~\citep{yang2025qwen3}, Llama-3.1-8B-Instruct and
Llama-3.3-70B-Instruct~\citep{grattafiori2024llama3},
Gemma-3-27B-IT~\citep{gemmateam2025gemma3}, and
Gemini-2.5-Flash-Lite~\citep{comanici2025gemini25}. The router refers to Workers only
by anonymous IDs with capability descriptions and never sees model names or prices.
Model identifiers and serving providers are listed in Appendix~\ref{app:models}.

\textbf{Baselines.}
We compare \ours with ten routers spanning all three execution paradigms.
Single-round baselines include Router-KNN and Router-MLP~\citep{hu2024routerbench},
RouterDC~\citep{chen2024routerdc}, and GraphRouter~\citep{feng2025graphrouter};
multi-round baselines include Prompt LLM and Router-KNN-MR following
\citet{zhang2025routerr1}, R2-Reasoner~\citep{shao2026r2reasoner}, and
Router-R1~\citep{zhang2025routerr1};
and agentic baselines include MasRouter~\citep{yue2025masrouter} and
GraphPlanner~\citep{feng2026graphplanner}.

\subsection{Performance Across Execution Paradigms}
\label{sec:main-results}

\input{tables/main}

\textbf{\ours substantially outperforms fixed-paradigm routers.}
Table~\ref{tab:main} compares \ours with single-round, multi-round, and agentic routing
baselines across the five evaluation domains. \ours reaches an average accuracy of
61.4\%, an 11.03\% relative improvement over the strongest baseline (55.3\%,
Router-R1). Its relative improvements over the strongest single-round, multi-round, and
agentic baselines are 26.08\%, 11.03\%, and 31.20\%, respectively, and it leads all
baselines in each of the five domains.

\textbf{Adaptive execution provides broad but task-dependent gains.}
\ours leads the strongest baseline in every evaluation domain, with relative
improvements ranging from 5.20\% on Recall to 12.52\% on Math. Importantly, the strongest
fixed-paradigm baseline varies across domains: multi-round routers perform best among
the baselines on Math, Code, and Knowledge, whereas single-round routers are strongest
on Reason and Recall. This variation suggests that no single execution
paradigm is uniformly preferable across tasks, motivating \ours's query-adaptive mode
selection. Section~\ref{sec:further-analysis} further examines how \ours distributes
queries across execution modes.

\subsection{Performance--Cost Trade-off}
\label{sec:cost-tradeoff}

\begin{wrapfigure}{R}{0.5\linewidth}
\vspace{-13pt}
\centering
\includegraphics[width=\linewidth]{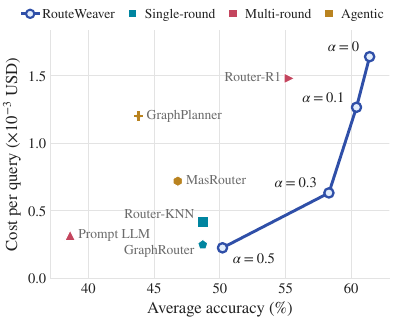}
\caption{Fixed-reference Worker inference cost per query against average accuracy. The
curve shows \ours trained with $\alpha \in \{0, 0.1, 0.3, 0.5\}$; markers show the two
strongest baselines of each execution paradigm in Table~\ref{tab:main}. The $\alpha = 0$ cost
is reconstructed from token usage.}
\label{fig:cost-tradeoff}
\vspace{-10pt}
\end{wrapfigure}

\textbf{Efficiency preferences allow \ours to operate at different accuracy--cost
points.}
The main results use $\alpha = 0$ and optimize task performance only, whereas different
deployments may require different balances between accuracy and cost. We therefore
evaluate \ours with $\alpha \in \{0, 0.1, 0.3, 0.5\}$, keeping all other settings
fixed; Figure~\ref{fig:cost-tradeoff} shows the resulting operating points. As $\alpha$
increases, Worker cost per query decreases monotonically, accompanied by a gradual
decrease in accuracy.
Relative to $\alpha = 0$, $\alpha = 0.1$ lowers cost by 23\% while losing 1.0 point of
accuracy, and $\alpha = 0.3$ lowers it by 61\% while losing 3.1 points; at
$\alpha = 0.5$, cost falls by 86\% while accuracy drops by 11.2 points. These results
illustrate different operating points that trade accuracy for efficiency under
different deployment preferences.

\textbf{Cost-aware routing provides different operating points under the same unified
policy.}
All operating points are obtained from the same unified routing framework by changing
only $\alpha$. Section~\ref{sec:further-analysis} analyzes how \ours distributes
execution modes across domains.

\subsection{Ablation of COMET-GRPO and Progressive Exploration}
\label{sec:ablation}

Table~\ref{tab:ablation} ablates the two learning components of \ours. Both variants
share the router initialization, Worker pool, training data, evaluation protocol, and
training budget of the full model, and all use $\alpha = 0$.

\textbf{w/o COMET-GRPO.}
Replacing COMET-GRPO with ordinary GRPO reduces average accuracy from 61.4\% to
58.8\% and shifts mode usage from .35/.41/.24 to .13/.82/.05, strongly concentrating
the policy on multi-round execution. The largest drop occurs on Knowledge (5.8
points). This suggests that decision-specific credit assignment helps preserve
differentiated mode selection rather than letting one execution mode dominate.

\input{tables/ablation}

\textbf{w/o progressive exploration.}
Removing the forced-to-free schedule causes a larger average drop, from 61.4\% to
55.6\%, with particularly large decreases on Code (9.6 points) and Reason (8.4
points). More strikingly, the learned policy collapses entirely to single-round
execution (1.00/.00/.00). This supports the role of progressive exploration in
maintaining cross-mode exploration before fully autonomous mode selection.
Notably, both ablated variants (58.8\% and 55.6\%) still exceed the strongest fixed-paradigm baseline (55.3\%, Router-R1).

\subsection{Routing Behavior and Generalization}
\label{sec:further-analysis}

\begin{wrapfigure}{r}{0.5\linewidth}
\vspace{-12pt}
\centering
\includegraphics[width=\linewidth]{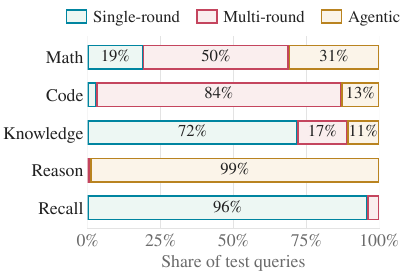}
\caption{Share of selected execution modes across evaluation domains for \ours
(\(\alpha=0\)). Each bar shows the proportion of test queries routed to single-round,
multi-round, and agentic execution.}
\label{fig:mode-usage}
\vspace{-10pt}
\end{wrapfigure}

\textbf{Routing behavior across domains.}
Figure~\ref{fig:mode-usage} reveals strongly task-dependent execution preferences. Math
uses all three modes, with 19\% of queries routed to single-round, 50\% to multi-round,
and 31\% to agentic execution. Code is dominated by multi-round routing (84\%), whereas
Knowledge primarily uses single-round execution (72\%). The contrast is strongest for
Reason and Recall: 99\% of reasoning queries are routed to agentic execution, while
96\% of recall queries use single-round routing. These results reveal substantially
different execution-mode preferences across task domains.

\textbf{Generalization to held-out benchmarks.}
To evaluate whether the learned routing policy transfers beyond the datasets used for
router training, we further test \ours on AIME~\citep{aimo2024validationaime,aime25}
for Math, GPQA-Diamond (GPQA-D)~\citep{rein2024gpqa} for Knowledge, and
BBH~\citep{suzgun-etal-2023-challenging} for Reason. We compare against one
representative baseline from each execution paradigm (Table~\ref{tab:generalization}).
\ours achieves the best accuracy on all three held-out benchmarks, reaching an average
of 53.7\%, an 18.28\% relative improvement over Router-R1 (45.4\%). The gains are
consistent across all three benchmarks, with improvements of 11.4, 7.1, and 6.4 points over Router-R1 on AIME, GPQA-D, and BBH, respectively. The learned execution behavior also varies across held-out benchmarks,
with agentic execution accounting for 75\% of AIME queries but only 12\% of BBH
queries. These results indicate that \ours's adaptive routing behavior extends to
unseen benchmarks within the same task domains.

\input{tables/ood}

%% file: tables/main.tex
\begin{table}[t]
\centering
\small
\caption{Accuracy (\%) on the evaluation benchmarks; \ours is trained with
$\alpha = 0$. Domain accuracy is the unweighted mean of its constituent dataset
accuracies; Avg is the unweighted mean of the five domain accuracies. Best results
are in \textbf{bold}; numbers in parentheses give the relative improvement of \ours
over the strongest baseline in each column.}
\label{tab:main}
\setlength{\abovetopsep}{4pt}
\setlength{\tabcolsep}{3pt}
\begin{tabularx}{\linewidth}{l*{6}{>{\centering\arraybackslash}X}}
\toprule
\textbf{Method} & \textbf{Math} & \textbf{Code} & \textbf{Knowledge} &
  \textbf{Reason} & \textbf{Recall} & \textbf{Avg} \\
\midrule
\multicolumn{7}{l}{\textbf{Single-round Router}} \\
\midrule
Router-KNN    & 61.7 & 34.8 & 38.3 & 68.3 & 40.4 & 48.7 \\
Router-MLP    & 58.7 & 33.2 & 40.0 & 59.2 & 34.2 & 45.1 \\
RouterDC      & 61.3 & 29.4 & 34.2 & 57.5 & 35.0 & 43.5 \\
GraphRouter   & 64.6 & 36.1 & 34.2 & 69.2 & 39.6 & 48.7 \\
\midrule
\multicolumn{7}{l}{\textbf{Multi-round Router}} \\
\midrule
Prompt LLM    & 45.8 & 33.7 & 29.2 & 54.2 & 30.0 & 38.6 \\
Router-KNN-MR & 40.8 & 33.2 & 30.8 & 48.3 & 29.6 & 36.6 \\
R2-Reasoner   & 45.8 & 30.7 & 28.3 & 50.0 & 23.3 & 35.6 \\
Router-R1     & 66.3 & 52.8 & 48.9 & 68.7 & 39.8 & 55.3 \\
\midrule
\multicolumn{7}{l}{\textbf{Agentic Router}} \\
\midrule
MasRouter     & 54.2 & 37.6 & 41.7 & 62.5 & 38.3 & 46.8 \\
GraphPlanner  & 47.5 & 48.3 & 30.0 & 54.2 & 39.2 & 43.8 \\
\midrule
\rowcolor{oursrow}
\ours         & \textbf{74.6}\imp{12.52} & \textbf{58.2}\imp{10.23}
              & \textbf{55.0}\imp{12.47} & \textbf{76.7}\imp{10.84}
              & \textbf{42.5}\imp{5.20}  & \textbf{61.4}\imp{11.03} \\
\bottomrule
\end{tabularx}

\end{table}

%% file: tables/ablation.tex

\begin{table}[t]
\centering
\small
\caption{Ablation of the two learning components, all trained with $\alpha = 0$ under
the same budget. Accuracy (\%) is reported by domain. Mode usage is reported as the
aggregate proportion of evaluation queries assigned to single-round (S), multi-round
(M), and agentic (A) execution.}
\label{tab:ablation}
\setlength{\abovetopsep}{4pt}
\setlength{\tabcolsep}{5.8pt}
\begin{tabular}{lccccccc}
\toprule
 & \multicolumn{6}{c}{\textbf{Accuracy (\%)}} & \textbf{Mode usage} \\
\cmidrule(lr){2-7} \cmidrule(lr){8-8}
\textbf{Variant} & \textbf{Math} & \textbf{Code} & \textbf{Knowledge} &
  \textbf{Reason} & \textbf{Recall} & \textbf{Avg} & \textbf{S / M / A} \\
\midrule
\rowcolor{oursrow}
\ours                        & 74.6 & 58.2 & 55.0 & 76.7 & 42.5 & 61.4 & .35 / .41 / .24 \\
w/o COMET-GRPO               & 72.3 & 57.2 & 49.2 & 74.1 & 41.2 & 58.8 & .13 / .82 / .05 \\
w/o progressive exploration  & 70.2 & 48.6 & 50.3 & 68.3 & 40.4 & 55.6 & 1.00 / .00 / .00 \\
\bottomrule
\end{tabular}
\end{table}

%% file: tables/ood.tex
\begin{table}[H]
\centering
\small
\caption{Accuracy (\%) on benchmarks held out from router training across three
domains: AIME (Math), GPQA-Diamond (Knowledge), and BBH (Reason). We compare one
representative router from each execution paradigm.}
\label{tab:generalization}
\setlength{\abovetopsep}{4pt}
\begin{tabular}{lcccc}
\toprule
\textbf{Method} & \textbf{AIME} & \textbf{GPQA-D} & \textbf{BBH} & \textbf{Avg} \\
\midrule
GraphRouter (single-round) & 23.5 & 49.5 & 24.7 & 32.6 \\
Router-R1 (multi-round)    & 45.6 & 62.1 & 28.6 & 45.4 \\
MasRouter (agentic)        & 28.1 & 43.8 & 28.2 & 33.4 \\
\midrule
\rowcolor{oursrow}
\ours                      & \textbf{57.0} & \textbf{69.2} & \textbf{35.0} & \textbf{53.7} \\
\bottomrule
\end{tabular}
\end{table}

%% file: sections/conclusion.tex

\section{Conclusion}
\label{sec:conclusion}

We presented \ours, a unified LLM routing framework that treats the execution paradigm
itself as a routing decision, jointly learning mode selection and within-mode
execution. By combining COMET-GRPO with progressive exploration, \ours learns when to
use single-round, multi-round, or agentic execution and how to route effectively
within the selected mode. Across five task domains, \ours consistently outperforms
fixed-paradigm routing baselines, exhibits task-dependent execution preferences, and
supports different performance--cost operating points. These results suggest that
effective LLM routing should adapt not only which models to invoke, but also how their
computation is organized. Extending this unified perspective to richer execution
structures and tool-augmented environments is an interesting direction for future
work.

%% file: sections/statements.tex

\subsection*{AI use statement}

In this work, we used generative AI tools to assist with implementation and debugging,
drafting and editing portions of the manuscript for clarity, and surveying and
summarizing relevant literature. All AI-assisted code and manuscript content were
reviewed by the authors, and AI-assisted code was inspected and tested before use. The
authors take full responsibility for the final content, claims, and artifacts
presented in this work.

\subsection*{Ethics statement}

This work studies routing among existing language models using established benchmarks
and does not involve human-subject experiments or the collection of private user data.
\ours inherits the limitations and potential harms of its underlying Worker models,
including biased, unsafe, or factually incorrect outputs, which may propagate across
multiple model calls. Our experiments are intended to evaluate routing methodology
rather than safety-critical deployment. Practical deployments should therefore apply
appropriate model-level safety mechanisms and comply with the licenses and usage
policies of the underlying models and datasets.

\subsection*{Reproducibility statement}

We provide the complete formulation and training objectives of \ours in
Section~\ref{sec:method}, and describe the datasets, router and Worker models, and
baselines in Section~\ref{sec:exp-setup}. The appendices further document the routing
serialization and execution constraints (Appendix~\ref{app:action-format}), dataset
construction and evaluation sets (Appendix~\ref{app:data}), task-specific evaluation
procedures (Appendix~\ref{app:eval}), model and serving configurations
(Appendix~\ref{app:models}), reward and cost definitions
(Appendix~\ref{app:training}), optimization details (Appendix~\ref{app:optim}),
and training hyperparameters, decoding settings, and
call budgets (Appendix~\ref{app:impl}). Together, these sections document the training
and evaluation setup used for \ours. Code is available at
\url{https://github.com/LaughKing/RouteWeaver}.



%% file: sections/appendix.tex

\clearpage
\startcontents[appendix]
\section*{Appendix Contents}
\vspace{-0.5em}
{\setlength{\parskip}{0pt}
\printcontents[appendix]{}{1}{\setcounter{tocdepth}{2}}}
\vspace{1em}

\section{Routing Interface}

\subsection{Action Format}
\label{app:action-format}

The router's output is written with the four elements illustrated in
Table~\ref{tab:representation}: one \texttt{mode} element, followed by \texttt{route}
elements interleaved with environment-inserted \texttt{observation} elements, and a
final \texttt{answer} element. Workers are named by their anonymous IDs.

\begin{table}[h]
\centering
\small
\caption{Schematic of the shared trajectory representation.}
\label{tab:representation}
\setlength{\tabcolsep}{5pt}
\begin{tabular}{@{}lll@{}}
\toprule
Element & Illustrative form & Produced by \\
\midrule
Mode selection &
  \texttt{\textless mode\textgreater agentic\textless /mode\textgreater} &
  Router \\
\addlinespace
Worker call &
  \begin{tabular}[t]{@{}l@{}}
    \texttt{\textless route model="worker\_2" role="verifier" refs="n1"\textgreater}\\
    \texttt{\ \ Check the proposed solution.\textless /route\textgreater}
  \end{tabular} &
  Router \\
\addlinespace
Returned output &
  \texttt{\textless observation id="n2"\textgreater \ldots
  \textless /observation\textgreater} &
  Environment \\
\addlinespace
Final response &
  \texttt{\textless answer\textgreater \ldots \textless /answer\textgreater} &
  Router \\
\bottomrule
\end{tabular}
\end{table}

Each Worker receives the sub-query written in its \texttt{route} element together
with the mode-specific context described below. A trajectory must follow the
grammar and the execution constraints of its mode:
\begin{itemize}
    \item \textbf{Single-round:} exactly one call
    \texttt{\textless route model="$w$"\textgreater}$s$\texttt{\textless /route\textgreater},
    followed by the final answer.
    \item \textbf{Multi-round:} two to four sequential calls of the same form, one per
    router turn; each call after the first also receives the previous observation.
    \item \textbf{Agentic:} calls of the form\\
    \hspace*{1em}\texttt{\textless route model="$w$" role="$r$" refs="$P$"\textgreater}$s$\texttt{\textless /route\textgreater},\\
    with $r \in \{$planner, executor, verifier, summarizer$\}$ and $P$ a
    comma-separated list of node IDs (empty for the first layer). Several routes in one
    router turn form a layer and run in parallel; a route may reference only outputs of
    earlier layers, which the environment inserts into that Worker's input. Each output
    returns as \texttt{\textless observation id="n$k$" role="$r$"\textgreater}. At most
    eight Worker calls and ten router action turns are allowed.
\end{itemize}
The first action that violates the grammar or an execution constraint of the mode
terminates the trajectory with rewards $(-1,-1)$ (Appendix~\ref{app:training}).
\subsection{Example Trajectories}

Figure~\ref{fig:examples} shows one abbreviated trajectory per mode from the
evaluation of the reported router.

\definecolor{modesingle}{RGB}{49,130,189}
\definecolor{modemulti}{RGB}{209,86,118}
\definecolor{modeagentic}{RGB}{224,151,38}
\tcbset{trajbox/.style n args={2}{
  colback=#2!4, colframe=#2!70, boxrule=0.5pt, arc=1.5pt,
  left=4pt, right=4pt, top=3pt, bottom=3pt, boxsep=1pt,
  fonttitle=\bfseries\footnotesize, coltitle=black, colbacktitle=#2!22,
  title=#1, before skip=4pt, after skip=4pt}}
\begin{figure}[h]
\centering
\begin{minipage}{0.97\linewidth}
\footnotesize
\begin{tcolorbox}[trajbox={Single-round}{modesingle}]
\begin{verbatim}
<mode>single</mode>
<route model="worker_3">If $x$ and $y$ are positive integers with
  $3^x 5^y = 225$,
  what is the value of $x + y$?</route>
<observation>...</observation>
<answer>4</answer>
\end{verbatim}
\end{tcolorbox}
\begin{tcolorbox}[trajbox={Multi-round}{modemulti}]
\begin{verbatim}
<mode>multi</mode>
<route model="worker_3">Calculate the number of three-digit integers
  from 100 to 999
  whose digit sum is between 7 and 11 inclusive.</route>
<observation>...</observation>
<route model="worker_3">Verify that the total number of three-digit
  integers with
  digit sum between 7 and 11 inclusive is 224, and confirm the last
    two digits
  are 24.</route>
<observation>...</observation>
<answer>24</answer>
\end{verbatim}
\end{tcolorbox}
\begin{tcolorbox}[trajbox={Agentic}{modeagentic}]
\begin{verbatim}
<mode>agentic</mode>
<route model="worker_3" role="planner" refs="">Split the problem
  into sub-tasks: ...
  </route>
<observation id="n1" role="planner">...</observation>
<route model="worker_3" role="executor" refs="n1">Compute the number
  of unit cubes
  with 0 gold faces and exactly 1 gold face for $n = 9$, ...</route>
<observation id="n2" role="executor">...</observation>
<route model="worker_3" role="verifier" refs="n1,n2">Verify that the
  formulas ...
  </route>
<observation id="n3" role="verifier">...</observation>
<route model="worker_3" role="summarizer" refs="n1,n2,n3">Reconcile
  the planner's
  decomposition, the executor's computation, and the verifier's
    audit ...</route>
<observation id="n4" role="summarizer">...</observation>
<answer>9</answer>
\end{verbatim}
\end{tcolorbox}
\end{minipage}
\caption{Abbreviated trajectories of the reported router, one per mode (Omni-MATH
evaluation questions). Observation bodies and parts of long sub-queries are elided
(\ldots), the routing-state messages that the environment inserts between agentic
layers are omitted, and line breaks inside routes are added for display.}
\label{fig:examples}
\end{figure}

\section{Experimental Setup}

\subsection{Datasets}
\label{app:data}

\paragraph{Training data.}
The router is trained on 3{,}200 queries from seven datasets
(Table~\ref{tab:train-data}). Each training batch contains 32 queries with fixed
domain quotas: 7 Math, 7 Code (4 APPS and 3 LiveCodeBench), 6 Knowledge, 7 Reason, and
5 Recall.

\begin{table}[h]
\centering
\small
\caption{Training data by domain.}
\label{tab:train-data}
\begin{tabular}{@{}l l r r@{}}
\toprule
Domain & Dataset & \#Queries & Per batch \\
\midrule
Math      & Omni-MATH                         & 700 & 7 \\
Code      & APPS, LiveCodeBench               & 400, 300 & 4, 3 \\
Knowledge & SuperGPQA                         & 600 & 6 \\
Reason    & SATBench                          & 700 & 7 \\
Recall    & TriviaQA, PopQA~\citep{mallen2023popqa} & 200, 300 & 5 \\
\midrule
Total     &                                   & 3{,}200 & 32 \\
\bottomrule
\end{tabular}
\end{table}

\paragraph{Evaluation data.}
Evaluation uses 1{,}060 test questions that do not overlap with the training
questions: 120 each from MATH, Omni-MATH, APPS, LiveCodeBench, SuperGPQA, SATBench,
TriviaQA, and SimpleQA-Verified, and 100 from TACO.

\subsection{Evaluation Protocol}
\label{app:eval}

\paragraph{Per-dataset scoring.}
Every question receives a binary score, and we report accuracy per dataset.
\begin{itemize}
    \item \textbf{MATH, Omni-MATH:} symbolic mathematical equivalence with the
    reference answer.
    \item \textbf{APPS:} the extracted program is executed in a sandbox on standard
    input and is correct if its output matches on all test cases.
    \item \textbf{LiveCodeBench, TACO:} extracted code is executed and is correct
    if it passes all retained test cases. At most 40 test cases are kept per problem;
    for LiveCodeBench, test cases are additionally kept in their original order only
    while their cumulative serialized size stays within 256\,KB (at least one is
    always kept). Both rules are deterministic, so the full official test suites are
    not always used.
    \item \textbf{SuperGPQA:} the last option letter in the response must match
    the reference exactly.
    \item \textbf{SATBench:} the last occurrence of SAT or UNSAT in the response must
    match the reference label.
    \item \textbf{SimpleQA-Verified:} normalized exact match, or the reference
    answer appears in the response as a whole-word match.
    \item \textbf{TriviaQA:} exact match with any reference alias after SQuAD-style
    normalization (lowercasing and removing punctuation, articles, and extra
    whitespace).
\end{itemize}

\paragraph{Aggregation.}
Domain accuracy is the unweighted mean of the accuracies of its constituent datasets,
and Avg is the unweighted mean of the five domain accuracies.

\subsection{Router and Worker Models}
\label{app:models}

\paragraph{Router.}
Qwen3-4B-Instruct-2507, with full-parameter fine-tuning. The KL reference policy is
the frozen initial router. Only the router is trained.

\paragraph{Workers.}
The six Workers are frozen and are exposed to the router under fixed anonymous IDs.
The router's system prompt lists each ID with a fixed capability description
(Table~\ref{tab:workers}) and never contains model names or prices. worker\_1 and
worker\_2 are deployed locally on two NVIDIA A100 GPUs, and worker\_3 to worker\_6 are
accessed through OpenRouter; the serving setup is the same in training and
evaluation.

\begin{table}[h]
\centering
\small
\caption{Worker pool and the capability descriptions shown to the router.}
\label{tab:workers}
\begin{tabular}{@{}l l p{0.52\linewidth}@{}}
\toprule
ID & Model & Capability description in the router prompt \\
\midrule
worker\_1 & Qwen3-8B &
  Dense language model with 8.2B parameters. Supports multilingual instruction
  following, dialogue, mathematical and logical reasoning, and code generation. \\
worker\_2 & Llama-3.1-8B-Instruct &
  Dense language model with 8B parameters. Instruction-tuned for multilingual
  dialogue, with capabilities in general reasoning, text generation, and coding. \\
worker\_3 & Qwen3-30B-A3B-Instruct-2507 &
  Mixture-of-experts language model with 30.5B total and 3.3B active parameters.
  Supports instruction following, text comprehension, logical reasoning, mathematics,
  science, and coding. \\
worker\_4 & Gemma-3-27B-IT &
  Dense language model in the 27B parameter class. Supports multilingual question
  answering, summarization, reasoning, mathematical problem solving, and code
  generation. \\
worker\_5 & Llama-3.3-70B-Instruct &
  Dense language model with 70B parameters. Instruction-tuned for multilingual
  dialogue, with capabilities in reasoning, instruction following, and text and code
  generation. \\
worker\_6 & Gemini-2.5-Flash-Lite &
  Sparse mixture-of-experts model with undisclosed parameter counts. Supports
  translation, classification, information extraction, coding, mathematics, and
  reasoning. \\
\bottomrule
\end{tabular}
\end{table}

\section{Training Details}

\subsection{Rewards and Cost Accounting}
\label{app:training}

This section completes the definitions of Section~\ref{sec:method}.

\paragraph{Rewards and excluded rollouts.}
Each trajectory receives the two rewards of Eq.~\ref{eq:reward}. Correctness is judged
by the dataset-specific scorers of Appendix~\ref{app:eval}. The first action that
violates the grammar
or an execution constraint of the mode (Appendix~\ref{app:action-format}) terminates
the trajectory, which receives $(-1,-1)$, remains in $\mathcal{V}$, and enters the
group statistics like any other rollout. A rollout affected by an infrastructure
failure, such as a Worker call returning no response, is not attributed to the router
and is not treated as terminal-invalid; it is excluded from $\mathcal{V}$ and hence
from all group statistics, the policy loss, and the KL term. When the efficiency reward is enabled
($\alpha > 0$), trajectories with incomplete cost records are excluded from
$\mathcal{V}$ in the same way, so they enter neither reward component nor the
corresponding updates.

\paragraph{Cost measure.}
$C(\tau)$ sums the inference cost of every Worker call in $\tau$, computed from the
call's input and output tokens at the fixed reference prices of
Table~\ref{tab:prices} (in USD per million tokens). The same frozen table is used for
the efficiency reward and for all reported costs. These prices serve only to measure
Worker inference cost; they are not the amounts actually paid, and the router's own
cost is not included. The normalization constant $C_{\mathrm{ref}} = 0.0070716653$
USD is the 95th percentile of the per-trajectory Worker cost over 1{,}024 rollouts
(128 training queries, 8 rollouts each) sampled from the step-25 cold-start
checkpoint with the training decoding settings, excluding only infrastructure failures
and trajectories with incomplete usage records. It is frozen for all runs, so that
$S_{\mathrm{cost}}(\tau) = 1/2$ when $C(\tau) = C_{\mathrm{ref}}$.

\begin{table}[h]
\centering
\small
\caption{Fixed reference prices (USD per million tokens).}
\label{tab:prices}
\begin{tabular}{@{}l l r r@{}}
\toprule
ID & Model & Input & Output \\
\midrule
worker\_1 & Qwen3-8B                    & 0.020 & 0.050 \\
worker\_2 & Llama-3.1-8B-Instruct       & 0.020 & 0.050 \\
worker\_3 & Qwen3-30B-A3B-Instruct-2507 & 0.100 & 0.300 \\
worker\_4 & Gemma-3-27B-IT              & 0.119 & 0.200 \\
worker\_5 & Llama-3.3-70B-Instruct      & 0.100 & 0.320 \\
worker\_6 & Gemini-2.5-Flash-Lite       & 0.100 & 0.400 \\
\bottomrule
\end{tabular}
\end{table}

\subsection{Optimization Details}
\label{app:optim}

\paragraph{Component-wise group statistics.}
For a reward component $c \in \{\mathrm{task},\mathrm{eff}\}$ with values $r_{i,c}$, a
group with retained rollouts $\mathcal{V}$ and mode subsets $\mathcal{V}_m$ has
\begin{equation}
    \mu_c = \frac{1}{|\mathcal{V}|} \sum_{i \in \mathcal{V}} r_{i,c},
    \qquad
    \mu_{m,c} = \frac{1}{|\mathcal{V}_m|} \sum_{i \in \mathcal{V}_m} r_{i,c},
    \qquad
    \sigma_c = \operatorname{std}\big(\{r_{i,c}\}_{i \in \mathcal{V}}\big),
\end{equation}
and advantages
\begin{equation}
    A_{i,c}^{\mathrm{GRPO}} = \frac{r_{i,c}-\mu_c}{\sigma_c+\epsilon},
    \qquad
    A_{i,c}^{\mathrm{mode}} = \frac{\mu_{m_i,c}-\mu_c}{\sigma_c+\epsilon},
    \qquad
    A_{i,c}^{\mathrm{inner}} = \frac{r_{i,c}-\mu_{m_i,c}}{\sigma_c+\epsilon}.
    \label{eq:component-adv}
\end{equation}
Here $\sigma_c$ is the sample (Bessel-corrected) standard deviation over all retained
rollouts of the group, not a per-mode one, and $\epsilon = 10^{-6}$.
The advantages used in training mix the components with the run's fixed weight
$\alpha$,
\begin{equation}
    A_i^{l} = (1-\alpha)\,A_{i,\mathrm{task}}^{l} + \alpha\,A_{i,\mathrm{eff}}^{l},
    \qquad l \in \{\mathrm{GRPO},\mathrm{mode},\mathrm{inner}\},
    \label{eq:component-mix}
\end{equation}
which extends Eq.~\ref{eq:adv-mix} to the ordinary GRPO advantage used by forced
groups. Each component is normalized by its own group statistics before mixing; the
rewards are never combined into a scalar and normalized jointly, and no further
batch-level normalization or whitening is applied.

\paragraph{Component-wise gating.}
Each component is gated separately: if $\sigma_c < 10^{-4}$, component $c$
contributes no advantage for that group. A group whose rollouts are all correct but
differ in cost therefore keeps its efficiency signal when $\alpha > 0$, although its
task signal is gated. A group whose rollouts are all excluded from $\mathcal{V}$
contributes no gradient.

\paragraph{Ordinary GRPO loss.}
With $S_i = S_i^{\mathrm{mode}} \cup S_i^{\mathrm{inner}}$ and
$N_{\mathrm{pol}} = \sum_i |S_i|$,
\begin{equation}
    \mathcal{L}_{\mathrm{GRPO}}^{\mathrm{global}} =
    \frac{1}{N_{\mathrm{pol}}} \sum_i \sum_{k \in S_i} \ell_{ik}(A_i^{\mathrm{GRPO}}).
    \label{eq:grpo-global}
\end{equation}
In cold start every mode prefix is injected, so $S_i^{\mathrm{mode}} = \varnothing$ and
$S_i$ contains only execution tokens, and only the task component is used,
$A_i^{\mathrm{GRPO}} = A_{i,\mathrm{task}}^{\mathrm{GRPO}}$.

\paragraph{Dual clipping.}
For negative advantages, the token loss of Eq.~\ref{eq:token-loss} is additionally
bounded as $\ell_{ik}(A) \leftarrow \min\{\ell_{ik}(A), -cA\}$ for $A < 0$, with
$c = 3$, which limits the loss of a negative-advantage token whose ratio has grown
large. The clipping ranges are $\eta_- = \eta_+ = 0.2$.

\paragraph{KL regularization.}
$\mathcal{L}_{\mathrm{KL}}$ in Eq.~\ref{eq:kl} is computed against the frozen initial
router, the original Qwen3-4B-Instruct-2507 rather than the cold-start checkpoint,
with coefficient $\beta$ (Appendix~\ref{app:impl}); $\kappa_{ik}$ is the low-variance
($k_3$) estimator, and the KL term enters the loss rather than the reward. It is
normalized by its own token count $N_{\mathrm{KL}}$ over $\Omega_{\mathrm{KL}}$, not by
the segment counts of Eq.~\ref{eq:pg-loss}. $\Omega_{\mathrm{KL}}$ uses the same
router-token mask as the policy loss, so rollouts excluded from $\mathcal{V}$ by
infrastructure failures are removed from both.

\paragraph{Mixed-batch token advantages.}
During the transition, the token advantage used in Eq.~\ref{eq:pg-loss} is
\begin{equation}
    \hat{A}_{ik} =
    \begin{cases}
        A_i^{\mathrm{mode}},  & k \in S_i^{\mathrm{mode}},\ \text{free group},\\[2pt]
        A_i^{\mathrm{inner}}, & k \in S_i^{\mathrm{inner}},\ \text{free group},\\[2pt]
        A_i^{\mathrm{GRPO}},  & k \in S_i^{\mathrm{inner}},\ \text{forced group},
    \end{cases}
    \label{eq:mixed-adv}
\end{equation}
where all advantages are the $\alpha$-mixed values of Eq.~\ref{eq:component-mix}. A
group is either entirely forced or entirely free. For forced groups the injected mode
tokens are masked, so $S_i^{\mathrm{mode}} = \varnothing$ and the execution tokens
receive $A_i^{\mathrm{GRPO}}$, which equals $A_i^{\mathrm{inner}}$ because all rollouts
of the group share one mode; these tokens count toward $N_{\mathrm{inner}}$.

\paragraph{Support-term scores.}
Let $s_\theta(m,x)$ be the summed log probability of the content tokens of candidate
mode $m$ for query $x$. The scores are normalized per query and averaged over the set
$\mathcal{U}$ of unique queries in the batch:
\begin{equation}
    \tilde{p}_\theta(m \mid x) =
    \frac{\exp s_\theta(m,x)}{\sum_{m' \in \mathcal{M}} \exp s_\theta(m',x)},
    \qquad
    \bar{p}_m = \frac{1}{|\mathcal{U}|} \sum_{x \in \mathcal{U}}
    \tilde{p}_\theta(m \mid x).
    \label{eq:support-scores}
\end{equation}
$\bar{p}_m$ is thus an average of normalized candidate-mode scores, not the frequency
with which $m$ is sampled, and the support term does not ensure that every mode is
sampled in every batch.

\paragraph{Degenerate groups.}
For each component, if a mode has only one retained rollout, then
$\mu_{m_i,c} = r_{i,c}$ and that rollout's inner advantage is zero; if all retained
rollouts select the same mode, then $\mu_{m_i,c} = \mu_c$ and all mode advantages are
zero. Gating is described above.

\paragraph{Relation to GRPO advantages.}
For each component,
$A_{i,c}^{\mathrm{mode}} + A_{i,c}^{\mathrm{inner}} = A_{i,c}^{\mathrm{GRPO}}$, and the
same identity holds for the mixed advantages. The mode and inner advantages supervise
disjoint token sets with separate normalization.

\subsection{Hyperparameters and Decoding}
\label{app:impl}

\paragraph{Training.}
We use a batch size of 32 queries, 8 rollouts per query, a learning rate of
$10^{-6}$, a KL coefficient of 0.001 against the frozen initial router, and zero
entropy regularization. Each run, for $\alpha \in \{0,0.1,0.3,0.5\}$, starts
independently from the same initial router at step 0, is trained for 90 steps, and is
evaluated at its final checkpoint. Steps 1--25 form the cold-start phase: forced modes, standard GRPO on the
task reward with the standard-deviation gate, global token reduction, and no support
term. After step 25 each run continues its own training trajectory, enabling the
mode/inner decomposition and segment-wise reduction and mixing the task and efficiency
advantages with its fixed $\alpha$; the KL reference remains the frozen initial
router. The forced ratio $\rho_b$ decreases linearly from 1 at data batch index 25 to
0 at batch index 75, and the support coefficient is $0.002\,\rho_b$. The support floor is $\delta = 0.05$ with $\lambda_0 = 0.002$,
the clipping ranges are $\eta_- = \eta_+ = 0.2$, and dual clipping uses $c = 3$.
The optimizer is AdamW with $\beta = (0.9, 0.999)$, $\epsilon = 10^{-8}$, and weight
decay 0.01, using a constant learning rate without warmup or decay and global
gradient-norm clipping at 1.0. Each step performs a single on-policy update over all
$32 \times 8 = 256$ trajectories of the batch, so $\pi_{\mathrm{old}}$ is the policy
that generated them. In the ablation without progressive exploration, $\rho_b = 0$ for
all $b$, so the support term ($\lambda_b = \lambda_0 \rho_b$) is inactive as well.

\paragraph{Decoding.}
Router, training: temperature 0.8, top-$p$ 0.95, top-$k$ disabled. Router, testing:
greedy decoding. Workers, training and testing: temperature 0.2, at most 4{,}096
output tokens; top-$p$ and top-$k$ are not set and follow provider defaults. Worker
requests carry a fixed seed (42), which providers may ignore, and disable
thinking/reasoning modes. Because Workers sample, system-level evaluation is not fully
deterministic. Each evaluation question receives one system rollout.

\paragraph{Call budgets and lengths.}
Identical in training and evaluation. Single-round: exactly one Worker call.
Multi-round: two to four sequential calls. Agentic: at most eight Worker calls and at
most ten router action turns. Prompts are
limited to 6{,}144 tokens, the multi-turn response budget to 12{,}288 tokens, the
mode-declaration turn to 64 tokens, and each observation to 2{,}048 tokens.